\documentclass[runningheads]{llncs}
\pdfoutput=1
\usepackage{graphicx}
\usepackage{amsmath,amssymb}
\usepackage{color}

\usepackage{subcaption}
\usepackage{multirow}

\usepackage{comment}
\usepackage{mathtools}
\usepackage{algorithm}
\usepackage{algpseudocode}
\usepackage{hyperref}
\hypersetup{
    colorlinks=true,
    linkcolor=blue,
    filecolor=magenta,      
    urlcolor=magenta,
}

\begin{document}

\pagestyle{headings}
\mainmatter

\title{Visually Guided Sound Source Separation using Cascaded Opponent Filter Network}

\titlerunning{Cascaded Opponent Filter Network}
%
\author{Lingyu Zhu \and
Esa Rahtu}
\authorrunning{Zhu and Rahtu}
%
\institute{Tampere University, Finland\\
\email{lingyu.zhu@tuni.fi} and \email{esa.rahtu@tuni.fi}}
\maketitle

\begin{abstract}
The objective of this paper is to recover the original component signals from a mixture audio with the aid of visual cues of the sound sources. Such task is usually referred as visually guided sound source separation. The proposed \textit{Cascaded Opponent Filter} (COF) framework consists of multiple stages, which recursively refine the source separation. A key element in COF is a novel opponent filter module that identifies and relocates residual components between sources. The system is guided by the appearance and motion of the source, and, for this purpose, we study different representations based on video frames, optical flows, dynamic images, and their combinations. Finally, we propose a \textit{Sound Source Location Masking} (SSLM) technique, which, together with COF, produces a pixel level mask of the source location. The entire system is trained end-to-end using a large set of unlabelled videos. We compare COF with recent baselines and obtain the state-of-the-art performance in three challenging datasets (\textit{MUSIC}, \textit{A-MUSIC}, and \textit{A-NATURAL}). Project page: \href{https://ly-zhu.github.io/cof-net}{https://ly-zhu.github.io/cof-net}.



\end{abstract}

\section{Introduction}

\begin{figure}[!tbp]
    \centering
    \includegraphics[width=0.85\textwidth,keepaspectratio]{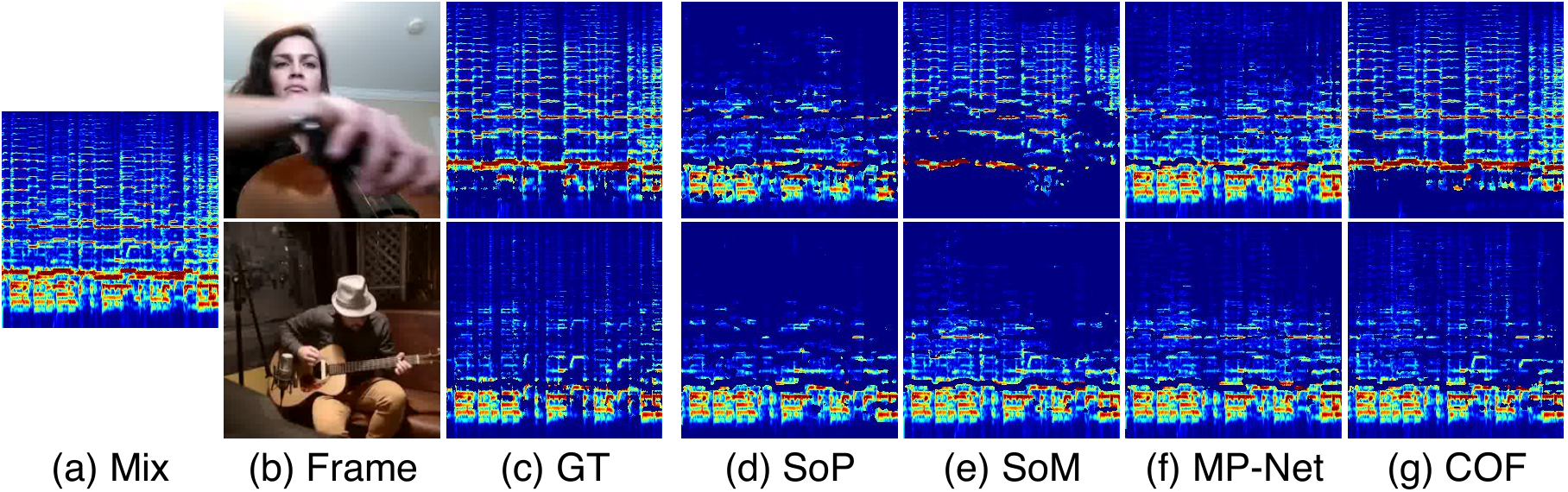}
    \caption{Visually guided sound source separation aims at splitting the input mixture (left) into component signals corresponding to the given visual cues (top and bottom row). The proposed COF approach results in better separation performance over the baseline methods SoP~\cite{zhao2018sound}, SoM~\cite{zhao2019sound}, and MP-Net~\cite{xu2019recursive} on MUSIC dataset~\cite{zhao2018sound}.
   }
    \label{fig:comp_vis}
\end{figure}

Sound source separation \cite{ghahramani1996factorial,roweis2001one,cichocki2009nonnegative,virtanen2007monaural} is a classical audio processing problem, where the objective is to recover original component signals from a given mixture audio. Well known example of such task is the cocktail party problem, where multiple people are talking simultaneously (e.g.~at a cocktail party) and the observer is attempting to follow one of the discussions. The general form of the problem is challenging and highly underdetermined. Fortunately, one is often able to leverage additional constraints from external cues, such as vision. For instance, the cocktail party problem turns more tractable by observing the lip movements of people \cite{ephrat2018looking}. Similar visual cues have also been applied in other sound separation tasks \cite{gao2018learning,owens2018audio,zhao2018sound,zhao2019sound,xu2019recursive,gao20192,gao2019co}. This type of problem setup is often referred as visually guided sound source separation (see e.g.~Fig.~\ref{fig:comp_vis}).

Besides separating the component signals from the mixture, one is often interested in identifying the source location. Such task would be intractable from a single audio channel, but could be approached using e.g.~microphone arrays~\cite{pertila2011closed}. Alternatively, the sound source location could be determined from the visual data \cite{tian2018audio,hu2019deep}, which is more often available.

This paper proposes a new approach for visually guided sound source separation and localisation. Our system (Fig. \ref{fig:overview}), referred as Cascaded Opponent Filter (COF), consists of an initial separation stage and one or more subsequent cascaded Opponent Filter (OF) modules (Fig. \ref{fig:OF}). The OF module utilises visual cues from \emph{all} videos to reconstruct each component audio. This is in contrast to most previous works (e.g.~\cite{zhao2018sound,zhao2019sound}), where the separation is done only based on corresponding video. The OF is very light containing only 17 parameters (in our case) and we show that it can greatly improve the sound separation performance over the recent single stage systems \cite{zhao2018sound,zhao2019sound} and recursive method \cite{xu2019recursive}.

Moreover, since motion is strongly correlated to sound formation \cite{zhao2019sound}, we build our system on both appearance and motion representations. To this end, we examine multiple options based on video frames, optical flows, dynamic images~\cite{bilen2016dynamic}, and their combinations. Finally, we introduce a Sound Source Location Masking (SSLM) network that, in conjunction with COF, is able to pin point pixel level segmentation of source locations. Qualitative results indicate sharper and more accurate results compared to the baselines \cite{zhao2018sound,zhao2019sound,xu2019recursive}. The entire system is trained using a self-supervised setup with large set of unlabelled videos. 

\section{Related Work}

\paragraph{\bf Cross-modal Learning from Audio and Vision} 
Aytar~\textit{et al.}~\cite{aytar2016soundnet} presented a method for learning joint audio-visual embeddings by minimizing the KL-divergence of their representations. Owens~\textit{et al.}~\cite{owens2016ambient} proposed a synchronization based cross-modal approach for visual representation learning. Arandjelovic~\textit{et al.}~\cite{arandjelovic2017look,arandjelovic2018objects} associated the learnt audio and visual embeddings by asking whether they originate from the same video. Nagrani~\textit{et al.}~\cite{nagrani2018seeing} learned to identify face and voice correspondences. More recent works, include transferring mono- to binaural audio using visual features~\cite{gao20192}, audio-video deep clustering~\cite{alwassel2019self}, talking face generation~\cite{zhou2019talking}, audio-driven 3D facial animation prediction~\cite{cudeiro2019capture}, and speech embedding disentanglements~\cite{nagrani2020disentangled}. Unlike these works, (visually guided) sound source separation aims at splitting the input audio into original components signals. 

\paragraph{\bf Video Sequence Representations} 
Most early works in video representations were largely based on direct extensions of the image based models \cite{laptev2008learning,wang2009evaluation,klaser2008spatio}.  More recently, these have been replaced by deep learning alternatives operating on stack of consecutive video frames. These works can be roughly divided into following categories: 1) 3D CNN applied on spatio-temporal video volume~\cite{tran2015learning}; 2) two-stream CNNs~\cite{simonyan2014two,carreira2017quo,zhan2019self} applied on video frames and separately computed optical flow frames; 3) LSTM~\cite{donahue2015long}, Graph CNN~\cite{wang2018videos} and attention clusters~\cite{long2018attention} based techniques; and 4) 2D CNN with the concept of dynamic image~\cite{bilen2016dynamic}. Since most of these methods are proposed for action recognition problem, it is unclear which representation would be best suited for self-supervised sound source separation. Therefore, this paper evaluates multiple options and discusses their pros and cons. 

\paragraph{\bf (Visually Guided) Sound Source Separation} The sound source separation task is extensively studied in the audio processing community. Early works were mainly based on probabilistic models~\cite{ghahramani1996factorial,roweis2001one,cichocki2009nonnegative,virtanen2007monaural}, while recent methods utilise deep learning architectures \cite{simpson2015deep,chandna2017monoaural,hershey2016deep,grais2018combining}. Despite of the substantial improvements, the pure audio based source separation remains a challenging task. At the same time, visually guided sound source separation has gained increasing attention. Ephrat~\textit{et al.}~\cite{ephrat2018looking} extracted face embeddings to facilitate speech separation. Similarly, Gao~\textit{et al.}~\cite{gao2019co,gao2018learning} utilised object detection and category information to guide source separation. Gan~\textit{et al.}~\cite{gan2020music} associated body and finger movements with audio signals by learning a keypoint-based structured representation. While impressive, these methods rely on the external knowledge of the video content (e.g. speaking faces, object types, or keypoints).

The works by Zhao~\textit{et al.}~\cite{zhao2018sound,zhao2019sound} and Xu~\textit{et al.}~\cite{xu2019recursive} are most related to ours. In \cite{zhao2018sound} the input spectrogram is split into components using U-Net~\cite{ronneberger2015u} architecture and the separated outputs are constructed as a linear combinations of these. The mixing coefficients are estimated by applying Dilated ResNet to the keyframes representing the sources. The subsequent work \cite{zhao2019sound} introduced motion features and improvements to the output spectrogram prediction. Both of these methods operate in a single stage manner directly predicting the final output. Alternatively, Xu~\textit{et al.}~\cite{xu2019recursive} proposed to separate sounds by recursively removing large energy components from the sound mixture. Our work explores multiple approaches to utilize the appearance and motion information to refine the sound source separation in multi-stages. The proposed Opponent Filter uses visual features of a sound source to look for incorrectly assigned sound components from opponent sources, resulting in accurate sound separation.

\paragraph{\bf Sound Source localization }

Early work by Hershey~\textit{et al.}~\cite{hershey2000audio} localised sound sources by modelling the audio-visual synchrony as a non-stationary Gaussian process. Barzelay~\textit{et al.}~\cite{barzelay2007harmony} applied cross-modal association and visual localization by temporal coincidences. Based on canonical correlations, Kidron~\textit{et al.}~\cite{kidron2005pixels} localized visual events associated with sound sources. Recently, Senocak~\textit{et al.}~\cite{senocak2018learning} learned to localize sound sources in visual scenes by transferring the sound-guided visual concepts to sound context vector. Arandjelovic~\textit{et al.}~\cite{arandjelovic2018objects} obtained locations by comparing visual and audio embeddings using a coarse grid. Class activation maps were used by \cite{owens2018learning,owens2018audio}. Gao~\textit{et al.}~\cite{gao2019co} localised potential sound sources via a separate object detector. Zhao~\textit{et al.}~\cite{zhao2018sound,zhao2019sound} and Xu~\textit{et al.}~\cite{xu2019recursive} visualise the sound sources by calculating the sound volume at each spatial location. In contrast to these methods, which either produce coarse sound location or rely on the external knowledge, we propose a self-supervised SSLM network to localise sound sources on a pixel level.

\section{Methods}
\label{sec:methods}

\begin{figure}[!tbp]
    \centering
    \includegraphics[width=0.9\textwidth,keepaspectratio]{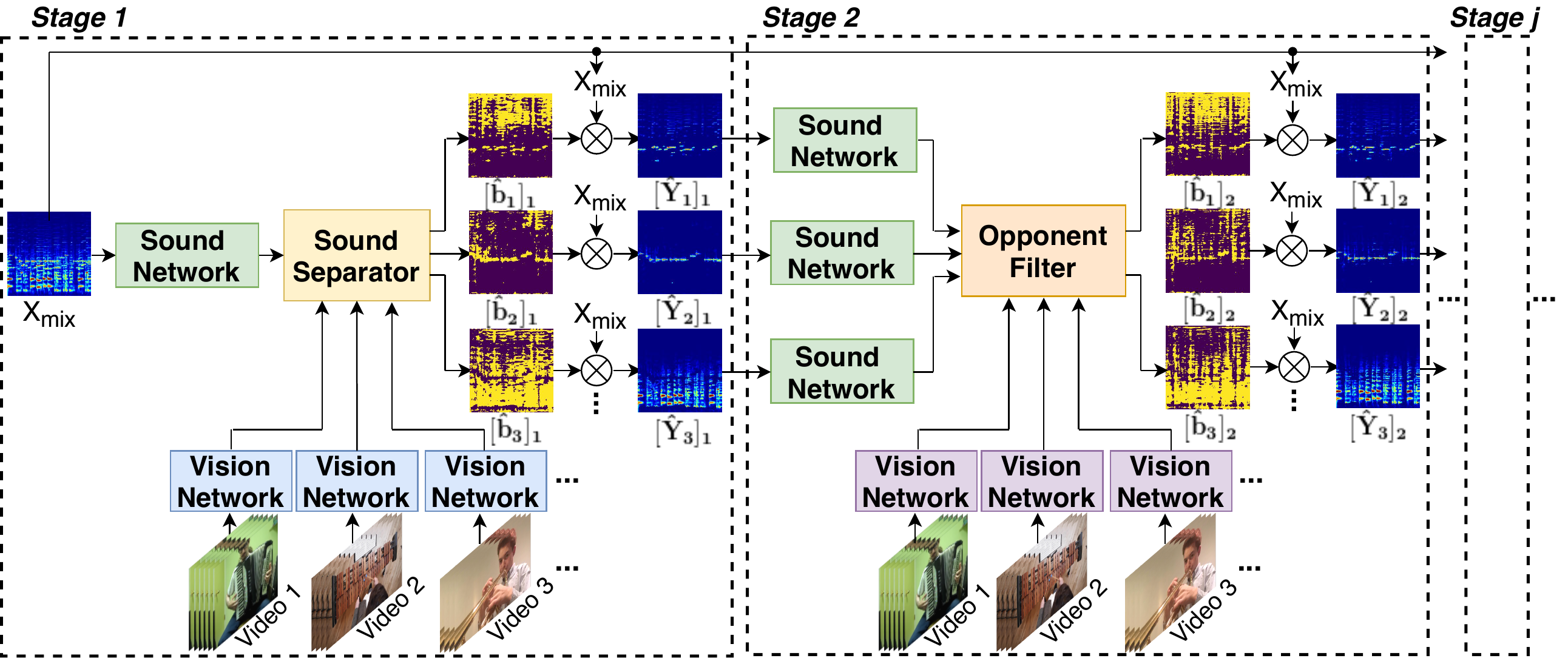}
    \caption{Architecture of the proposed Cascaded Opponent Filter (COF) network. COF operates in multiple stages: In the first stage, visual representations (vision network) and sound features (sound network) are passed to the sound separator and further produce a binary mask (Eq. (1), (2)) for each output source. Stage two refines the separation result using the opponent filter (OF) module guided by the visual cues. Later stages are identical to second stage with OF module.}
    \label{fig:overview}
\end{figure}

This section describes the proposed visually guided sound source separation method. We start with a short overview and then continue to detailed descriptions of each component.

\subsection{Overview}

The inputs to our system consist of a mixture audio (e.g.~band playing) and a set of videos, each representing one component of the mixture (e.g.~person playing a guitar). The objective of the system is to recover the component sound signals corresponding to each video sequence. Fig.~\ref{fig:overview} illustrates an overview of the approach. Note that the audio signals are represented as spectrograms, which are obtained from the audio stream using Short-term Fourier transform (STFT). 

The proposed system consists of multiple cascaded stages. The first stage contains three components: 1) a sound network that splits the input spectrogram into a set of feature maps; 2) a vision network that converts the input video sequences into compact representations; and 3) a sound separator that produces spectrum masks (not shown in Fig. \ref{fig:overview}) of the component audios (one per video) based on the outputs of the sound and vision networks.

The second stage contains similar sound and vision networks as the first one (internal details may differ). However, instead of the sound separator, the second stage contains a special Opponent Filter (OF) module, which enhances the separation result by transferring sound components between the sources. The output of the filter is passed to the next stage or used as the final output. The following stages are identical to the second one and, for this reason, we refer our method as cascaded opponent filter (COF) network. The final component audios are produced by applying the inverse STFT to the predicted component spectrograms. 

In addition, we propose a new Sound Source Location Masking (SSLM) network (not shown in Fig.~\ref{fig:overview}) that indicates the pixels with highest impact on the sound source separation (i.e.~source location). The entire network is trained in end-to-end fashion using artificially generated examples. That is, we take two or more videos and create an artificial mixture by summing the corresponding audio tracks. The created mixture and video frames are provided to the system, which then has to reproduce the original component audios. In the following sections, we will present each component with more details and provide the learning objective used in the training phase.

\subsection{Vision Network}
\label{sec:vision}

\begin{figure}[!tbp]
    \centering
    \includegraphics[width=\textwidth,keepaspectratio]{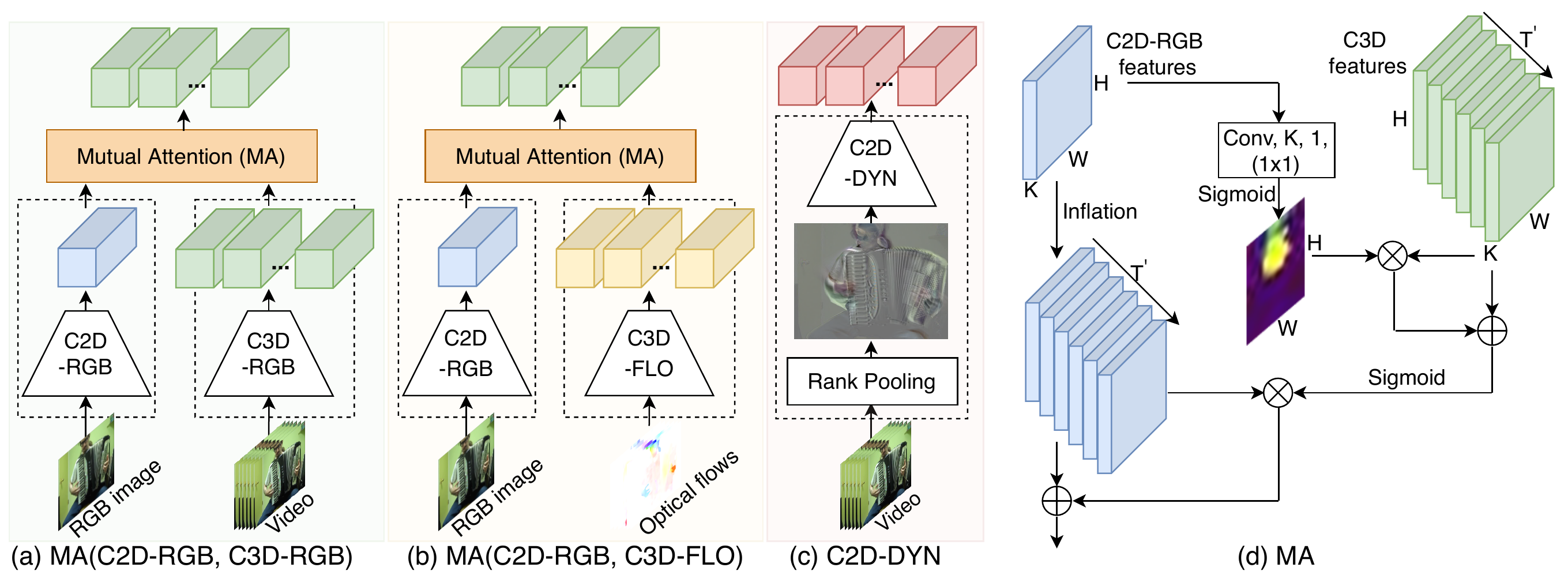}
    \caption{Architecture of (a) MA(C2D-RGB, C3D-RGB), (b) MA(C2D-RGB, C3D-FLO), (c) C2D-DYN, and (d) MA: Mutual Attention module.}
    \label{fig:sSep0_f48}
\end{figure}

The vision network aims at converting the input video sequence (or keyframe) into a compact representation that contains the necessary information of the sound source. Sometimes already a pure appearance of the source (e.g.~instrument type) might be sufficient, but, in most cases, the motions are vital cues to facilitate the source separation (e.g.~hand motion, mouth motion, etc.). The appropriate representation may have high model/computation complexity and, to seek for a balance between computational complexity and performance, we study several visual representation options. The models are introduced in the following and the detailed network architectures are provided in the supplementary material. In all cases, we assume that the input video sequence is of size $\textit{3}\times\textit{16H}\times\textit{16W}$ and has $\textit{T}$ frames. 

The first option, referred as \textbf{C2D-RGB}, is a pure appearance based representation. This is obtained by applying a dilated ResNet18~\cite{he2016deep} to a single keyframe extracted from the sequence. More specifically, given an input RGB image of size $\textit{3}\times\textit{16H}\times\textit{16W}$, the C2D-RGB produces a representation of size $\textit{K}\times\textit{H}\times\textit{W}$. Dynamic image~\cite{bilen2016dynamic} is a compact representation, which summarises the appearance and motion of the entire video sequence into a single RGB image by rank pooling the original pixel data. The second option, referred as \textbf{C2D-DYN}, first converts the input video into a dynamic image (size $\textit{3}\times\textit{16H}\times\textit{16W}$) and then applies a dilated ResNet18~\cite{he2016deep} to produce a representation of size $\textit{K}\times\textit{H}\times\textit{W}$. Fig.~\ref{fig:sSep0_f48}c illustrates C2D-DYN option. 

The third option, referred as \textbf{C3D-RGB}, applies 3D CNN to extract the appearance and motion information from the sequence simultaneously. C3D-RGB uses 3D version of ResNet18 and produces a representation of size ${\textit{T}}^{'}\times\textit{K}\times\textit{H}\times\textit{W}$. The optical flow~\cite{simonyan2014two,sun2018pwc,hui2018liteflownet} explicitly describes the motion between the video frames. The fourth option, referred as \textbf{C3D-FLO}, first estimates the optical flow between the consecutive video frames using LiteFlowNet~\cite{hui2018liteflownet}, and then applies 3D ResNet18 to the obtained flow sequence. C3D-FLO produces a representation of size ${\textit{T}}^{'}\times\textit{K}\times\textit{H}\times\textit{W}$.

In addition, following the recent work \cite{carreira2017quo} in action recognition, we propose a set of two stream options by combining pairs of C2D-RGB, C3D-RGB, and C3D-FLO representations using Mutual Attention (MA) module. The module is depicted in Fig.~\ref{fig:sSep0_f48}d. It enhances the sound source relevant motions and eliminates motion irrelevant appearance by giving the mutual attention mechanism. Finally, we receive the mutual attentive features of dimension ${\textit{T}}^{'}\times\textit{K}\times\textit{H}\times\textit{W}$ from the two-stream structures, which are referred to as \textbf{MA(C2D-RGB, C3D-RGB)} and \textbf{MA(C2D-RGB, C3D-FLO)}. Fig.~\ref{fig:sSep0_f48}a and \ref{fig:sSep0_f48}b illustrate these options. We omit the model of two 3D streams MA(C3D-RGB, C3D-FLO) due to large size of the resulting model.

\subsection{Sound Network}
\label{sec:sound}

The sound network splits the input audio spectrogram into a set of feature maps. The network is implemented using U-Net~\cite{ronneberger2015u} architecture and it converts the input spectrogram of size $\textit{HS}\times\textit{WS}$ into an output of size $\textit{HS}\times\textit{WS}\times\textit{K}$. Note that the number of created feature maps \textit{K} is equal to the visual feature dimension \textit{K} in the previous section. At the first stage, the input to the sound network is the original mixture spectrogram $\textit{X}_{\textit{mix}}$, while in later stages, the sound network operates on the current estimates of the component spectrograms. This allows stages to focus on different details of the spectrogram. In the following, we will denote the $k$th feature map, produced by the sound network for an input spectrogram $\textit{X}$, as $S(X)_{k}$.

\vspace{-3pt}
\subsection{Sound Separator}
\label{sec:SS}

The sound separator combines the visual representations with the sound network output and produces an estimate of the component signals. First, we apply global max pooling operation over the spatial dimensions ($\textit{H}\times\textit{W}$) of the visual representation. For 3D CNN based options, we further apply max pooling layer along the temporal dimension ($\textit{T}^{'}$). As a result, we obtain a feature vector $\mathbf{z}$ with $\textit{K}$ elements. We combine $\mathbf{z}$ with sound network output using a linear combination to predict the spectrum masks $g$ as Eq.~(\ref{eq:1}).

\begin{equation}
    \centering
    \label{eq:1}
        \textit{g}(\mathbf{z}, X) = \sum^K_{k=1} \alpha_{k} \, \mathbf{z_{k}} * S(X)_{k} + \beta, 
\end{equation}
where $\alpha_{\textit{k}}$ and $\beta$ are learnable weight parameters, $\mathbf{z}_k$ is the $k$th element of visual vector $\mathbf{z}$, and $S(X)_{k}$ is the $k$th sound network feature map for a spectrogram $\textit{X}$.

\subsection{Opponent Filter Module}
\label{sec:OF}

\begin{figure}[!tbp]
    \centering
    \includegraphics[width=0.9\textwidth,keepaspectratio]{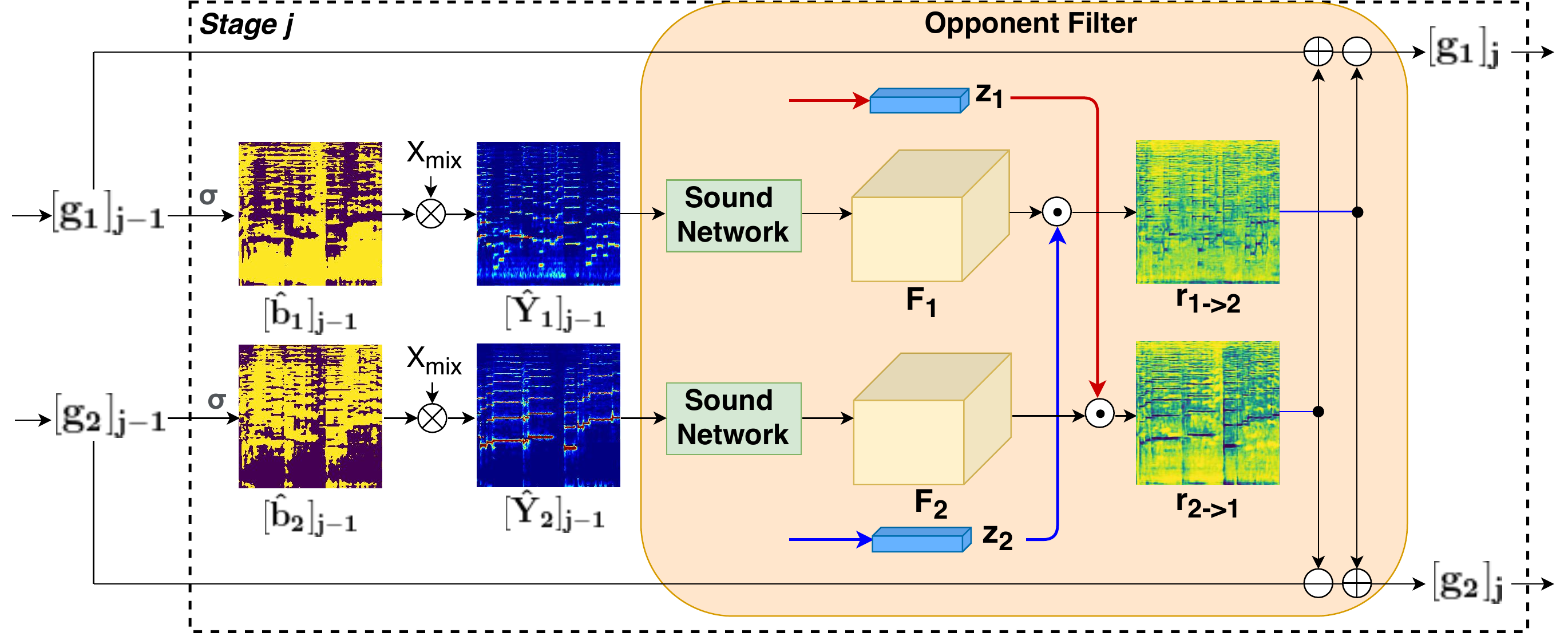}
    \caption{An illustration of the Opponent Filter (OF) module at stage $j$ in the case of two sound sources. The input consists of the visual representation $\mathbf{z}$ and the previous spectrum mask $[g]_{j-1}$, for both sources. First, we obtain the spectrograms $\hat{Y}$ for both sources from the spectrum masks (Eq.~(\ref{eq:2})). Second, the spectrograms are turned into feature maps $F$ with the sound network (Sec.~\ref{sec:sound}). Third, the visual representation $\mathbf{z}_2$ and the feature map $F_1$ are used to identify components from the source 1 that should belong to the source 2 ($r_{1->2}$ in figure). The spectrum masks are updated accordingly by subtracting from $[g_1]_{j-1}$ and adding to $[g_2]_{j-1}$. Similar operation is done for the source 2. Finally, the updated spectrum masks $[g_1]_{j}$ and $[g_2]_{j}$ are passed to the next stage.}
    \label{fig:OF}
\end{figure}

The structure of the Opponent Filter (OF) module is illustrated in Fig.~\ref{fig:OF} in the case of two sound sources. The main idea in the OF is to use visual representation of the source $n$ to identify spectrum components from the source $m$ that should belong to source $n$ but are currently assigned to $m$. These are then transferred from source $m$ to $n$. The motivation behind the construction is to utilise all visual representations $z_1,\ldots,z_N$ to determine each component audio, instead of using only the corresponding one. This is in contrast to the previous works SoP~\cite{zhao2018sound}, SoM~\cite{zhao2019sound} (and approximately for MP-Net~\cite{xu2019recursive}), where the output for each source is determined solely by the same visual input. Our approach leads to more efficient use of the visual cues, which is reflected by the performance improvements shown in the experiments (see Sec.~\ref{sec:exp_OF}). Moreover, in our case $(K=16)$, the selected architecture requires only 17 parameters per source pair, which makes it very light and efficient to learn. The OF module is used in all but the first stage of the COF.

More specifically, the OF module takes the visual representation $\mathbf{z}$ and the previous spectrum mask $[g]_{j-1}$ for each sound source as an input. Firstly, the spectrum masks are converted to the spectrograms $\hat{Y}$ as
\begin{equation}
    \centering
    \label{eq:2}
        \hat{b} = th( \sigma ({g})), \quad \hat{Y} = \hat{b} \otimes X_{\textit{mix}}
\end{equation}
where $\sigma$ denotes the sigmoid function, \textit{th} represents the thresholding operation with value 0.5, and $\otimes$ is the element-wise product. In other words, we first map $g$ into a binary mask $\hat{b}$, and then produce the estimate of the output component spectrogram as an element-wise multiplication between the binary mask $\hat{b}$ and the original mixture spectrogram $X_{\textit{mix}}$. $g$ and $\hat{Y}$ are provided for the upcoming stage as inputs (or used as the final output). We will denote the outputs corresponding to $n$th video at stage $j$ as $[{g}_n]_{j}$, $[\hat{b}_n]_{j}$, and $[\hat{Y}_n]_{j}$. The obtained spectograms are passed to the sound network (see Sec.~\ref{sec:sound}), which converts them to feature maps of size $\textit{HS}\times\textit{WS}\times\textit{K}$ denoted by $F_n$ for the source $n$. 

Secondly, the OF module takes one source at a time, referred using index $n\in[1,N]$, and iterates over the remaining sources $m\in\{[1,N] | m\not=n \}$. N is the number of sources in sound mixture. For each pair $(n,m)$ the filter determines a component of source $m$ that should be reassigned to the source $n$ as 
\begin{equation}
    \centering
    \label{eq:3}
        {r}_{m->n} = \sum^K_{k=1} \alpha_{k} \mathbf{z_{n,k}} * F_{m,k} + \beta
\end{equation}
where $\mathbf{z}_{n,k}$ is the $k$th element of visual representation of sound n. $F_{m,k}$ is the $k$th sound network feature maps of sound m. The ${r}_{m->n}$ denotes the residual spectrum components identified from source $m$ that should belong to source $n$ but are currently assigned to $m$. The obtained component will be subtracted from the spectrum mask $[{g}_m]_{j-1}$ and added to $[{g}_n]_{j-1}$ as follows
\begin{align}
    \centering
    \label{eq:4}
        [{g}_m]_{j} &= [{g}_m]_{j-1} \ominus {r}_{m->n} \\
        [{g}_n]_{j} &= [{g}_n]_{j-1} \oplus {r}_{m->n} 
\end{align}
where the $[g_n]_{j}$ is the spectrum mask (Eq.~(\ref{eq:1})) of $n$th video in stage $j$, ${r}_{m->n}$ is the residual spectrum components from sound m to sound n. $\oplus$ and $\ominus$ denote the element-wise sum and subtraction, respectively.

The overall process can be summarized in the following Algorithm 1,
\begin{algorithm} 
  \caption{Algorithm of Opponent Filter (OF) module}\label{alg:OF}
  \begin{algorithmic}[1]
          \For{n = 1 \dots N} 
            \For{m = 1 \dots N}
                \If{\texttt{$n \neq m$}}
                    \State ${r}_{m->n} = \sum^K_{k=1} \alpha_{k} \mathbf{z_{n,k}} * F_{m,k} + \beta $ \Comment{obtain $r_{m->n}$ }
                    \State $[{g}_m]_{j}\gets [{g}_m]_{j-1}\ominus r_{m->n} $ \Comment{subtract $r_{m->n}$ from $[{g}_m]_{j-1}$ }
                    \State $[{g}_n]_{j}\gets [{g}_n]_{j-1}\oplus r_{m->n} $ \Comment{add $r_{m->n}$ to $[{g}_n]_{j-1}$ }
                \EndIf
            \EndFor
          \EndFor
      \State \textbf{return} $[{g}]_{j}$ \Comment{return $[{g}]_{j}$ of all the sound sources}
  \end{algorithmic}
\end{algorithm}

\subsection{Learning Objective}
The model parameters are optimised with respect to the binary cross entropy (BCE) loss that is evaluated between the predicted and ground truth masks over all stages. More specifically, 
\begin{equation}
\centering
    \label{eq:5}
        \mathcal{L}_{\textit{sep}} = \sum^J_{j=1} r_j \, \textit{BCE}([\hat{b}]_j, b_{gt})
\end{equation}
where $r_j$ is a weight parameter, $[\hat{b}]_j$ is the predicted binary mask, $b_{gt}$ is the ground truth mask (determined by whether the target sound is the dominant component in the mixture), and $J$ is the total number of stages.

\subsection{Sound Source Location Masking Network}
\label{sec:localization}

\begin{figure}[!tbp]
    \centering
    \includegraphics[width=\textwidth,keepaspectratio]{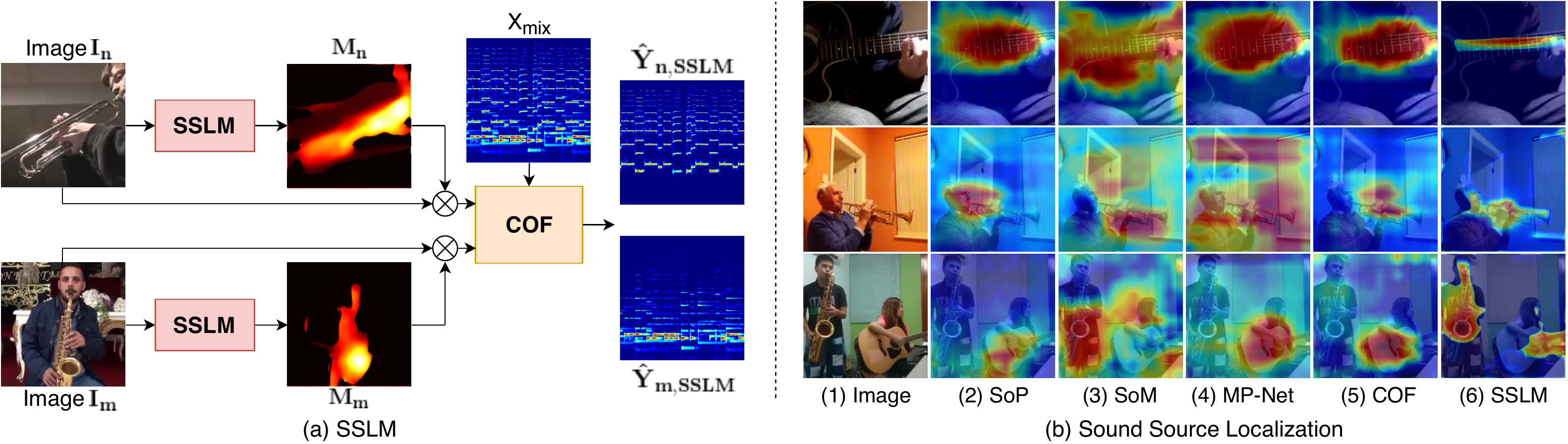}
    \caption{(a) The diagram of the Sound Source Location Masking (SSLM) network and (b) Visualizing sound source location of our methods in comparison with baseline models SoP~\cite{zhao2018sound}, SoM~\cite{zhao2019sound}, and MP-Net~\cite{xu2019recursive} on MUSIC dataset.}
    \label{fig:sslm}
\end{figure}

The objective of the Sound Source Location Masking (SSLM) network is to identify a minimum set of input pixels, for which the COF network would produce almost identical output as for the entire image. In practice, we follow the ideas presented in~\cite{hu2019visualization}, and build an auxiliary network to estimate a sound source location mask that is applied to the input RGB frames. The SSLM is trained together with the overall model. The input video frames are first passed through the SSLM component which outputs a weighted location mask [0,1] having same spatial size as the input frame. The input video frames are multiplied element-wise with the mask, and the result is passed to the COF model. We illustrate the overall structure of the SSLM in Fig.~\ref{fig:sslm}a. The SSLM network is implemented using a dilated residual network (DRN)~\cite{yu2017dilated} pre-trained on ImageNet~\cite{deng2009imagenet},  with three up-projection blocks~\cite{laina2016deeper} followed by a $3\times3$ convolution layer. The final optimisation is done by minimising the following loss function,
\begin{equation}
\centering
    \label{eq:6}
        \mathcal{L} = \sum^J_{j=1} r_j \, l_{\textit{diff}}([\hat{b}_{\textit{SSLM}}]_j, [\hat{b}]_j) + \lambda \frac{1}{q} \parallel \textit{SSLM}(I)\parallel_1,
\end{equation}
where $l_{\textit{diff}}$ denotes the difference between the $[\hat{b}_{\textit{SSLM}}]_j$ and $[\hat{b}]_j$ by $L_1$ norm, $[\hat{b}_{\textit{SSLM}}]_j$ is the output sound separation mask obtained using only selected pixels, $[\hat{b}]_j$ is the output separation mask for the original image, $r_j$ and $\lambda$ are weight parameters. The $\lambda \frac{1}{q} \parallel \textit{SSLM}(I)\parallel_1$ norm produces a location mask with only small number of non-zero values. q is the total number of pixels of the $\textit{SSLM}(I)$.

\section{Experiments}
\label{Sec:result}

We evaluate the proposed approach using Multimodel Sources of Instrument Combinations (MUSIC)~\cite{zhao2018sound} dataset, and two sub-sets of AudioSet~\cite{gemmeke2017audio}: A-MUSIC and A-NATURAL. The proposed model is trained using artificial examples, generated by adding audio signals from two of more training videos. The performance of the final sound source separation is measured in terms of standard metrics: Signal to Distortion Ratio (SDR), Signal to Interference Ratio (SIR), and Signal to Artifact Ratio (SAR). Higher is better for all metrics\footnote{Note that SDR and SIR scores measure the separation accuracy, SAR captures only the absence of artifacts (and hence can be high even if separation is poor)}.

\subsection{Datasets and Implementation Details}
\label{sec:dataset}

\subsubsection{MUSIC}
The Multimodel Sources of Instrument Combinations (MUSIC)~\cite{zhao2018sound} dataset is a relatively small, but has high quality. Most of the video frames are well aligned with the audio signals and have little off-screen noise. Part of the original MUSIC dataset is no longer available in YouTube (10\% missing at the time of writing). In order to keep dataset size, we replaced the missing entries with similar YouTube videos. Baseline methods (e.g., SoP~\cite{zhao2018sound}) in original paper split the dataset into 500 training and 130 validation videos, and report the performances on the validation set (train/test split are not published). Instead, we follow the standard practice of reporting the performance on a separate hold-out test set. For this purpose, we randomly split the dataset into 400 training, 100 validation, and 130 test videos. This leads to 20\% less training videos compared to~\cite{zhao2018sound}. All tested methods are trained and evaluated with the same data and pre-processing steps (see below). 

\subsubsection{A-MUSIC and A-NATURAL}
AudioSet consists of an expanding ontology of 632 audio event classes and is a collection of over 2 million 10-second sound clips drawn from YouTube videos. Many of the AudioSet videos have limited quality and sometimes the visual content might be uncorrelated to the audio track. A-MUSIC dataset is a trimmed musical instrument dataset from AudioSet. It has around 25k videos spanning ten instrumental categories. A-NATURAL dataset is a trimmed natural sound dataset from AudioSet. It contains around 10k videos which cover 10 categories of natural sounds. We split both the A-MUSIC and A-NATURAL dataset samples to 80\%, 10\%, and 10\% as train, validation and test set. More details of datasets are discussed in the supplementary material.

\subsubsection{Implementation Details}
We sub-sample each audio signals at 11kHz and randomly crop an audio clip of 6 seconds for training. A Time-Frequency (T-F) spectrogram of size $512\times256$ is obtained by applying STFT, with a Hanning window size of 1022 and a hop length of 256, to the input sound clip. We further re-sample this spectrogram to a T-F representation of size $256\times256$ on a log-frequency scale. We extract video frames at 8fps and give a single RGB image to the C2D-RGB model, $T=48$ frames to C2D-DYN and all the discussed C3D models. Further implementation details are provided in the supplementary material. 

\setlength{\tabcolsep}{4pt}
\begin{table}[t]
    \centering
    \caption{The sound separation results of the proposed COF network, conditioning on appearance cues, on MUSIC test dataset}
    \label{table:OF}
    \begin{tabular}{llll}
        \hline\noalign{\smallskip}
        Models & SDR & SIR & SAR\\
        \noalign{\smallskip}
        \hline
        \noalign{\smallskip}
        COF - 1 stage & 5.38 & 11.00 & 9.77 \\
        COF$_{\textit{addition}}$ - 2 stages & 6.29 & 11.83 & 10.21 \\
        COF$_{\textit{subtraction}}$ - 2 stages & 6.30 & 12.61 & 10.13 \\
        COF - 2 stages & $\mathbf{8.25}$ & $\mathbf{14.24}$ & $\mathbf{12.02}$ \\
        \hline
    \end{tabular}
\end{table}
\setlength{\tabcolsep}{1.4pt}

\subsection{Opponent Filter}
\label{sec:exp_OF}

In this section, we assess the performance of the OF module. For simplicity we use only the appearance based features (C2D-RGB) in all stages. The baseline is provided by the basic single stage version denoted as COF - 1 stage, which does not contain the opponent filter module. The results provided in Table~\ref{table:OF} indicate a clear improvement from the OF stages. 

In addition, we evaluate the impact of the ``addition'' and ``subtraction'' branches in the OF module. To this end, we implement two versions COF$_{addition}$ and COF$_{subtraction}$, which include only the ``addition'' (Eq. (5)) or ``subtraction'' (Eq. (4)) operation in the OF, respectively. The corresponding results in Table~\ref{table:OF} indicate that both versions obtain similar performance which is between the baseline and the full model. We conclude that both operations are essential part of the OF module and contribute equally to the sound separation result.

\subsection{Visual Representations}
\label{sec:VR}

\setlength{\tabcolsep}{4pt}
\begin{table}[t]
    \centering
    \caption{The sound separation results with COF, conditioning on different visual cues, on the MUSIC test dataset. Table contains three blocks: 1) single-stage COF associated with visual cues predicted from MA-RGB, MA-FLO, and C2D-DYN; 2) two-stage extension of the models in the first block; 3) two-stage COF with C2D-RGB at stage 1 and C3D-RGB, C3D-FLO, or C2D-DYN at stage 2}
    \label{table:vm}
    \begin{tabular}{lllll}
        \hline\noalign{\smallskip}
        & Models & SDR & SIR & SAR \\
        \noalign{\smallskip}
        \hline
        \noalign{\smallskip}
        \multirow{3}{*}{1} &
        COF(MA-RGB) & 6.68 & 12.24 & 10.63 \\
        & COF(MA-FLO) & 5.84 & 11.39 & 10.27 \\
        & COF(C2D-DYN) & 6.37 & 11.75 & 10.79 \\
        \noalign{\smallskip}
        \hline
        \noalign{\smallskip}
        \multirow{3}{*}{2} &
        COF(MA-RGB, MA-RGB) & 8.78 & 15.07 & 12.10 \\
        & COF(MA-FLO, MA-FLO) & 8.71 & 15.07 & 11.83 \\
        & COF(C2D-DYN, C2D-DYN) & 8.95 & 15.03 & 12.07 \\
        \noalign{\smallskip}
        \hline
        \noalign{\smallskip}
        \multirow{3}{*}{3} &
        COF(C2D-RGB, C3D-RGB) & 8.97 & 15.06 & \bf{12.53} \\
        & COF(C2D-RGB, C3D-FLO) & 9.04 & 15.28 & 12.24 \\
        & COF(C2D-RGB, C2D-DYN) & \bf{9.17} & \bf{15.32} & 12.37 \\
        \hline
        \end{tabular}
\end{table}
\setlength{\tabcolsep}{2pt}

We firstly separate sounds by implementing a single stage network, associating with the appearance and motion cues discussed in Sec.~\ref{sec:vision}. We denote the MA(C2D-RGB, C3D-RGB) and MA(C2D-RGB, C3D-FLO) as MA-RGB and MA-FLO in Table~\ref{table:vm}. As is shown in the block 1 of Table~\ref{table:vm}, the results with appearance and motion cues clearly surpass the network with only appearance cues from C2D-RGB in Table~\ref{table:OF}, which proposes that the motion representation is important for the sound separation quality. Block 2 shows the performance of how the visual information separates sounds in a two-stage manner. Explicitly, we replace the vision network at each stage in Fig.~\ref{fig:overview} with MA-RGB, MA-FLO, and C2D-DYN. Block 1 and 2 report that the three two-stage networks obtain similar performance and outperform their single-stage counterparts from block 1 with a large margin.

Finally, we evaluate an option where the first stage utilises only appearance based option and the second stage applies motion cues. In practice, we combine C2D-RGB with C3D-RGB, C3D-FLO, or C2D-DYN. The results in block 3 of Table~\ref{table:vm}, indicate that this combination obtains similar or even better performance than the options where motion information was provided for both stages. We conclude that the appearance information is enough to facilitate coarse separation at first stage. The motion information is only needed at the later stages to provide higher separation quality. It is worth noting that the COF(C2D-RGB, C2D-DYN) combination has less computation complexity and better performance compared to the 3D CNN alternatives. \textbf{Therefore, we apply C2D-RGB for the 1st stage and C2D-DYN for the later stages for all the remaining experiments}.

\subsection{Comparison with the State-of-the-Art}
\label{sec:SOTA}

\setlength{\tabcolsep}{4pt}
\begin{table}[!tbp]
    \centering
    \caption{Sound separation performance with 2 and 3 stages COF models compared with three recent baselines SoP~\cite{zhao2018sound}, SoM~\cite{zhao2019sound}, and MP-Net~\cite{xu2019recursive}, on MUSIC, A-MUSIC, and A-NATURAL datasets. The top 2 results are bolded.}
    \label{table:audioset}
    \begin{tabular}{llll|lll|lll}
        \hline\noalign{\smallskip}
        \multirow{2}{*}{Models $\backslash$ Datasets} &
            \multicolumn{3}{c}{MUSIC} &
            \multicolumn{3}{c}{A-MUSIC} &
            \multicolumn{3}{c}{A-NATURAL}\\
        \noalign{\smallskip}
        & SDR & SIR & SAR \quad & SDR & SIR & SAR \quad & SDR & SIR & SAR \\
        \noalign{\smallskip}
        \hline
        \noalign{\smallskip}
        SoP & 5.38 & 11.00 & 9.77 & 2.05 & 5.36 & 10.69 & 2.83 & 7.24 & 8.51\\
        SoM & 4.83 & 11.04 & 8.67 & 2.56 & 5.98 & 8.80 & 2.56 & 7.69 & 8.02\\
        MP-Net & 5.71 & 11.36 & 10.45 & 2.34 & 5.27 & \textbf{11.27} & 3.20 & 8.17 & 8.68\\
        COF - 2 stages & \textbf{9.17} & \textbf{15.32} & \textbf{12.37} & \textbf{3.31} & \textbf{7.08} & 10.74 & \textbf{4.00} & \textbf{8.85} & \textbf{8.70}\\
        COF - 3 stages & \textbf{10.07} & \textbf{16.69} & \textbf{13.02} & \textbf{5.42} & \textbf{9.47} & \textbf{10.94} & \textbf{4.10} & \textbf{8.60} & \textbf{10.58}\\
        \hline
    \end{tabular}
\end{table}
\setlength{\tabcolsep}{4pt}

\begin{figure}[!tbp]
    \centering
    \includegraphics[width=0.95\textwidth,keepaspectratio]{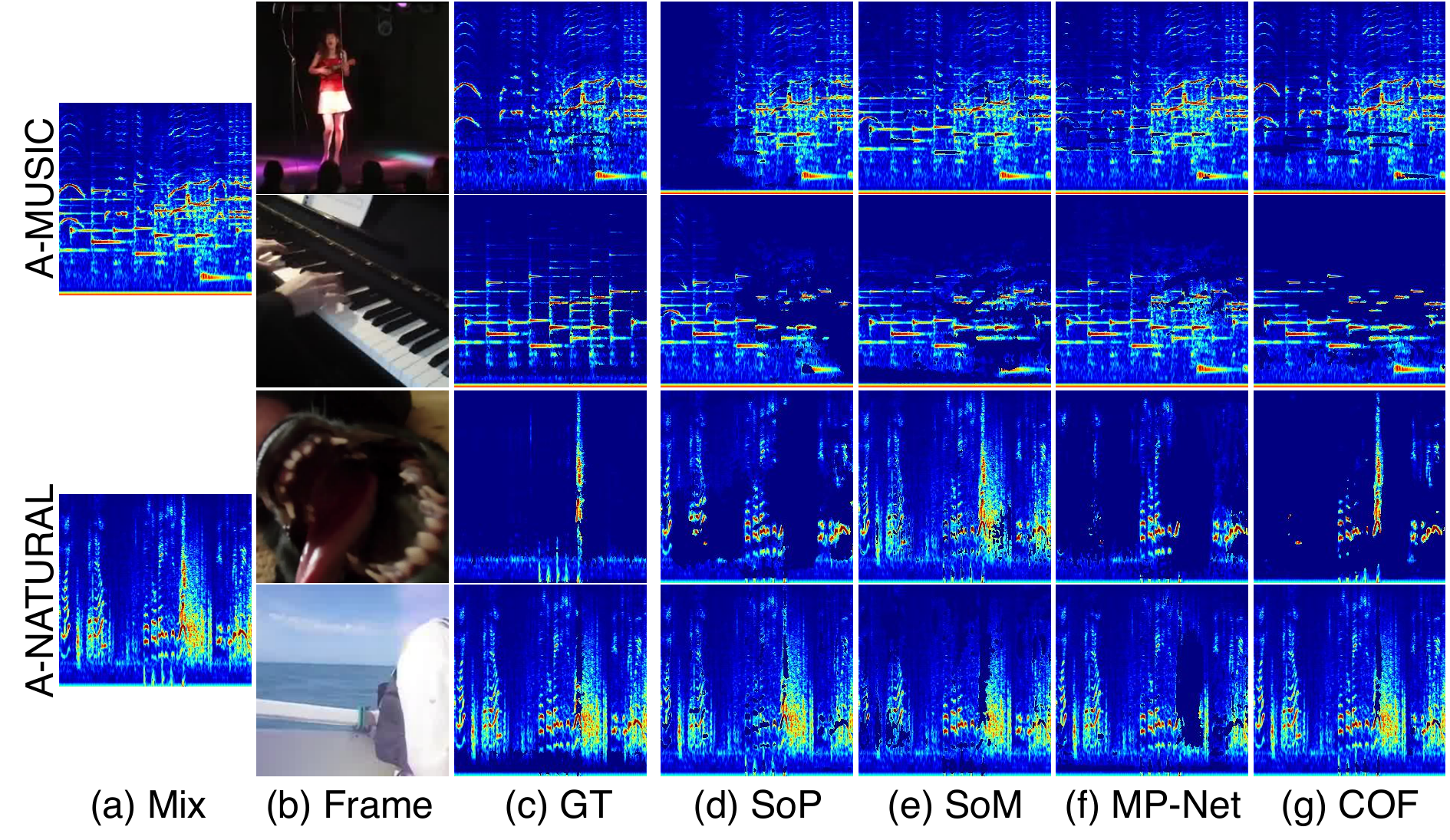}
    \caption{Visualizing sound source separation of our 2-stage COF model on A-MUSIC and A-NATURAL datasets, in comparison with baseline methods SoP~\cite{zhao2018sound}, SoM~\cite{zhao2019sound}, and MP-Net~\cite{xu2019recursive}.}
    \label{fig:sep_vis}
\end{figure}

We compare the 2-stage and 3-stage of the proposed COF model with three recent baseline methods SoP~\cite{zhao2018sound}, SoM~\cite{zhao2019sound}, and MP-Net~\cite{xu2019recursive}. For SoP we use the publicly available implementation from the original authors. For SoM and MP-Net we use our own implementations since there were no publicly available versions. The corresponding results for MUSIC\footnote{We note that due to the differences in the dataset and evaluation protocol (see Sec. 4.1.) the absolute results differ from those reported in \cite{zhao2018sound} and \cite{zhao2019sound} for MUSIC.}, A-MUSIC, and A-NATURAL datasets are provided in Table~\ref{table:audioset}, Fig.~\ref{fig:comp_vis}, and Fig.~\ref{fig:sep_vis}. The quantitative results indicate that our model outperforms the baselines with a large margin across all three datasets. 

\paragraph{Increasing the number of stages:} We observe that the computational cost increases approximately linearly with respect to the number of stages. The performance generally improves until it reaches a plateau. COF with 2, 3, 4, and 5 stages obtain SDRs of 9.17, 10.07, 10.12, and 10.32 on MUSIC dataset, respectively. The performance plateaus at 3 stages, which led to a compromise at this point.

\paragraph{Mixture of three sources:} We assess the COF model using a mixture of three sound sources from the MUSIC dataset. In this case, the two-stage model obtains SDR: 3.33, SIR: 10.32, and SAR: 6.70 which are clearly higher than SDR: 1.30, SIR: 8.66, and SAR: 5.73 obtained with MP-Net~\cite{xu2019recursive} that is particularly designed for the multi-source case. As discussed in Sec.~\ref{sec:OF}, the computational cost of COF scales approximately linearly with the number of sources. 

\subsection{Visualizing Sound Source Locations}
\label{sec:loc}

We compare the sound source localizing capability of our best two-stage model with state-of-the-art methods in Fig.~\ref{fig:sslm}b. Columns (2)-(5) display the sound energy distributions of spatial location in heatmaps on input frame during inference. COF produces precise associations between visual representation and separated sounds, though columns (5) is just the visualization from the first stage of COF. As we know, the spatial features from ConvNet usually have small resolution ($14\times14$ pixels in this work). Thus, the final visualized location is generally coarse after up-sampling the heatmap to the resolution of the input image. Differently, our proposed SSLM learns to predict a pixel-level sound source location mask, as shown in column (6), which precisely localizes sound sources and preserves high quality of sound separation. Further examples are provided in the supplementary material.

\section{Conclusions}

We proposed an innovative framework of visually guided Cascaded Opponent Filter (COF) network to recursively refine sound separation with visual cues of sound sources. The proposed Opponent Filter (OF) module uses visual features of a sound source to look for incorrectly assigned sound components from opponent sources, resulting in accurate sound separation. For this purpose, we studied different visual representations based on video frames, optical flows, dynamic images, and their combinations. Moreover, we introduced a Sound Source Location Making (SSLM) network, together with COF, to precisely localize sound sources. We have performed extensive evaluations on our proposed methods and obtained state-of-the-art performance on challenging datasets.

\clearpage
%
%
\bibliographystyle{splncs04}
\bibliography{ms}

\appendix
\section{Supplementary Material}

\subsection{Sound Source Separation for Instrument Combinations}

\begin{figure}[h]
    \centering
    \includegraphics[width=0.8\textwidth,keepaspectratio]{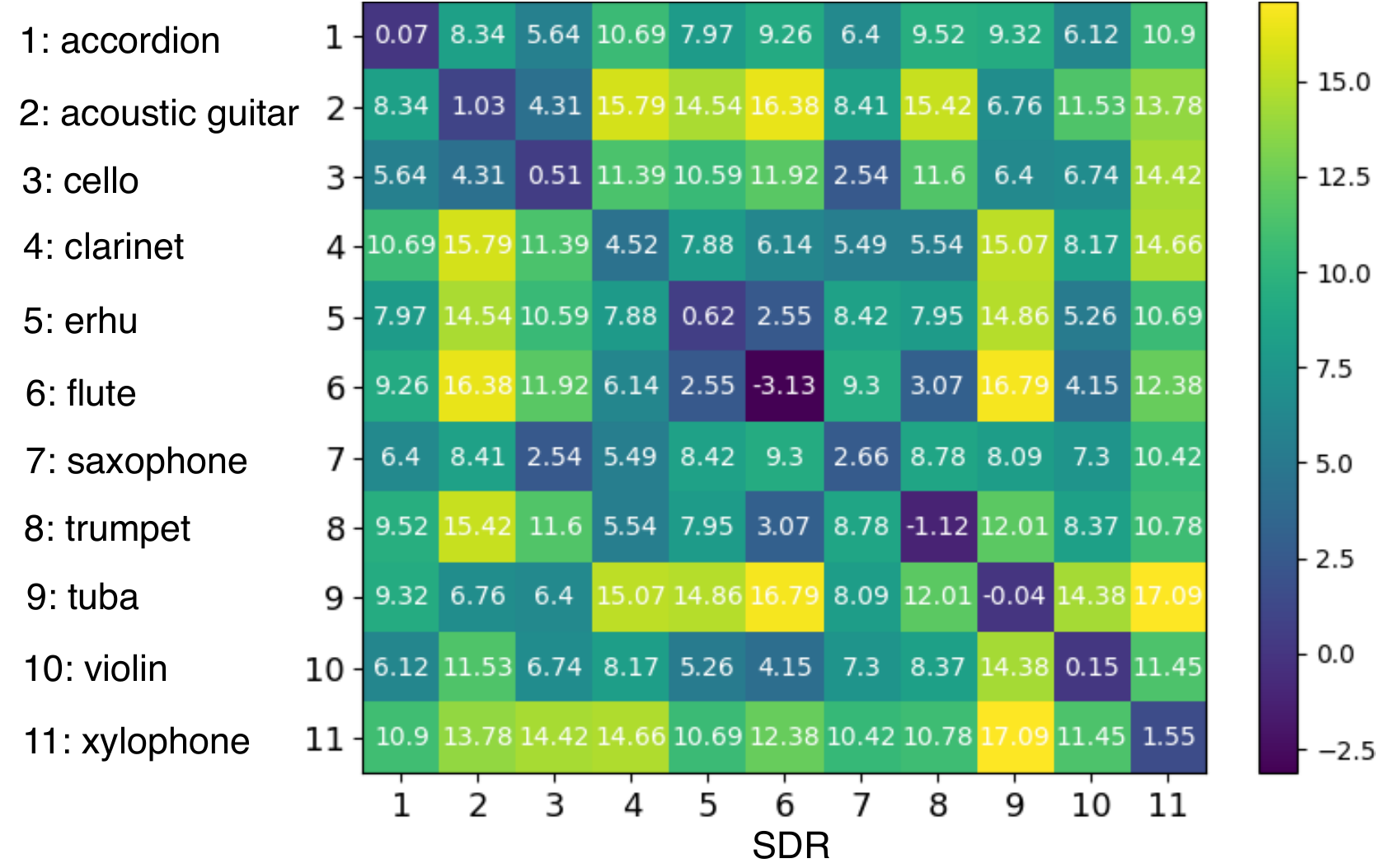}
    \caption{The sound source separation performance for different mixtures of instruments in MUSIC dataset. The results are shown in terms of SDR. The diagonals and off-diagonals represent the results of separating instrumental combinations of same and different categories respectively. The higher value of SDR represents the better performance of sound source separation, e.g. \textit{acoustic guitar} and \textit{flute} (2 and 6), \textit{flute} and \textit{tuba} (6 and 9), \textit{tuba} and \textit{xylophone} (9 and 11). The SDR values on the diagonals clearly indicate the hardest case of separating sounds between two instruments from the same category, e.g. \textit{flute} (6 and 6).}
    \label{fig:sep_matrix_2}
\end{figure}

In this section, we present sound source separation performance for different instrument mixtures using MUSIC dataset. Fig.~\ref{fig:sep_matrix_2} illustrates the results in terms of SDR in a matrix form. The diagonals represent the results of separating instruments of same categories (e.g.~two guitars), and the off-diagonals are combinations from different categories (e.g.~guitar and violin). The higher value of SDR represents the better performance of sound source separation, e.g. \textit{acoustic guitar} and \textit{flute} (2 and 6), \textit{flute} and \textit{tuba} (6 and 9), \textit{tuba} and \textit{xylophone} (9 and 11). The SDR values on the diagonals clearly indicate that separating sounds between two instruments from the same category is the hardest case. In particular, separating the mixture of two flutes is challenging. One reason might be the small amount of motion related to playing flute.

\subsection{Sound Source Localization Examples}

\begin{figure}[!tbp]
    \centering
    \includegraphics[width=0.85\textwidth,keepaspectratio]{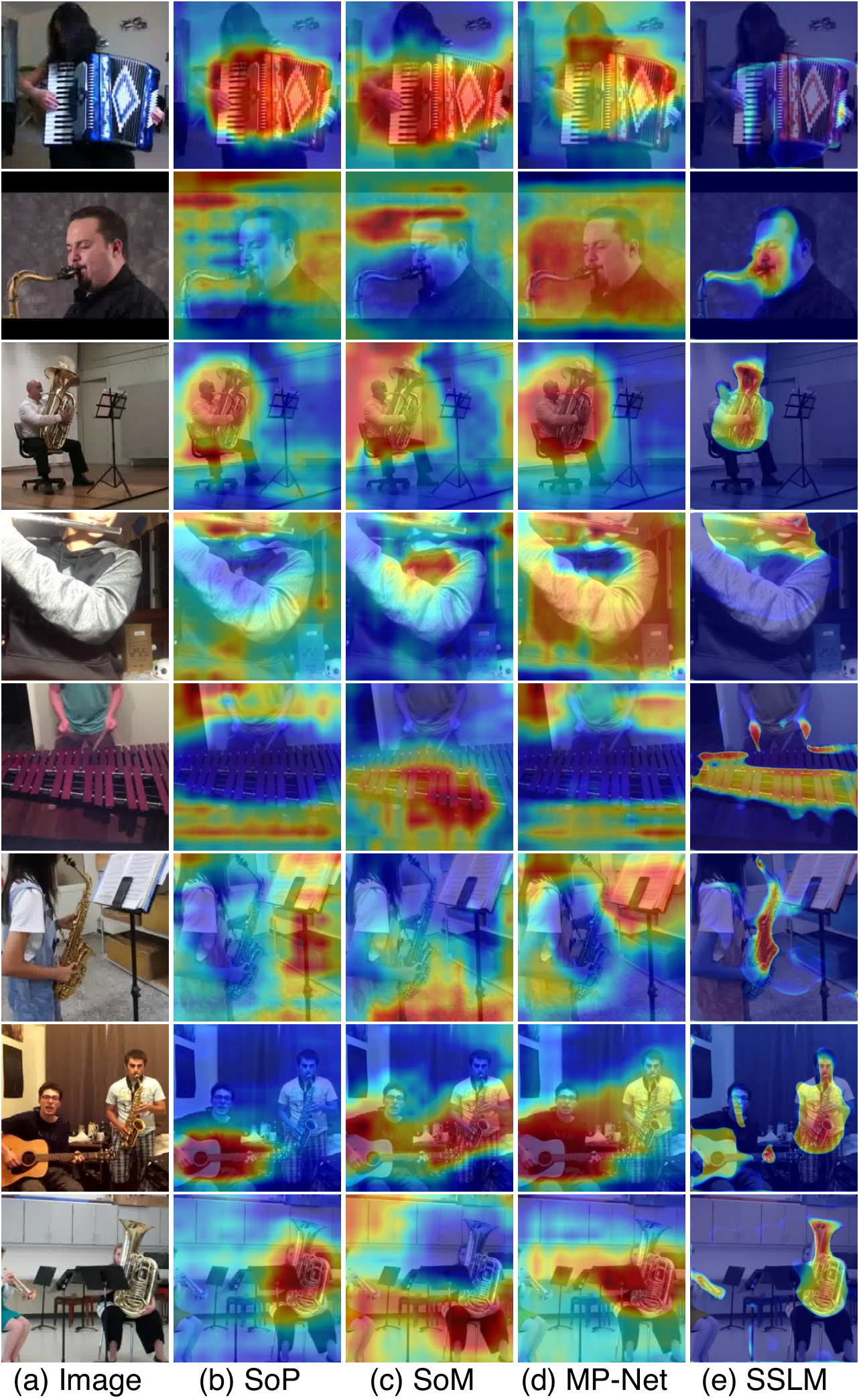}
    \caption{Visualizing sound source location of our proposed SSLM network in comparison with baseline methods SoP, SoM, and MP-Net on MUSIC dataset.}
    \label{fig:loc_music_vis}
\end{figure}

\begin{figure}[!tbp]
    \centering
    \includegraphics[width=0.85\textwidth,keepaspectratio]{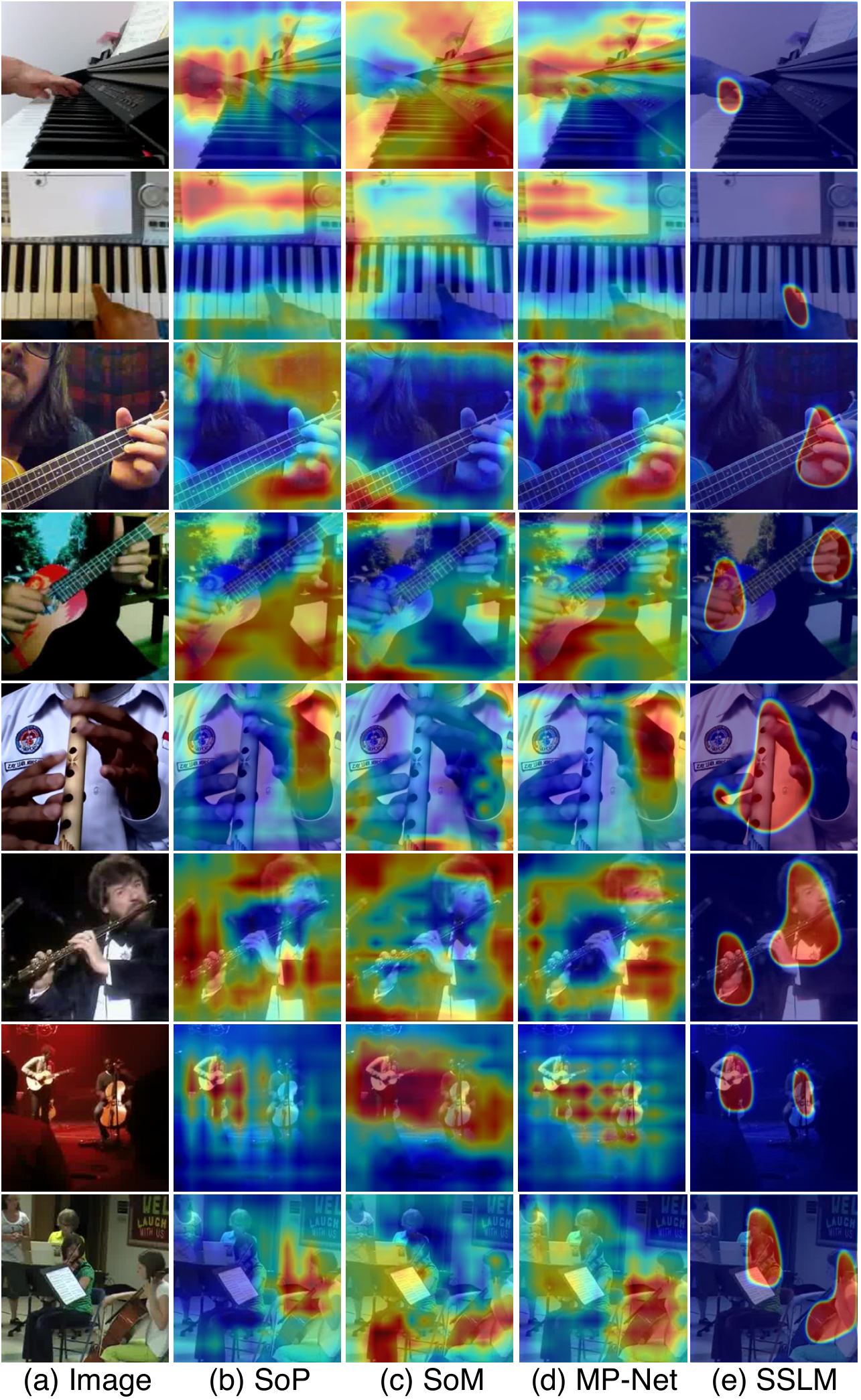}
    \caption{Visualizing sound source location of our proposed SSLM network in comparison with baseline methods SoP, SoM, and MP-Net on A-MUSIC dataset.}
    \label{fig:loc_amusic_vis}
\end{figure}

\begin{figure}[!tbp]
    \centering
    \includegraphics[width=0.85\textwidth,keepaspectratio]{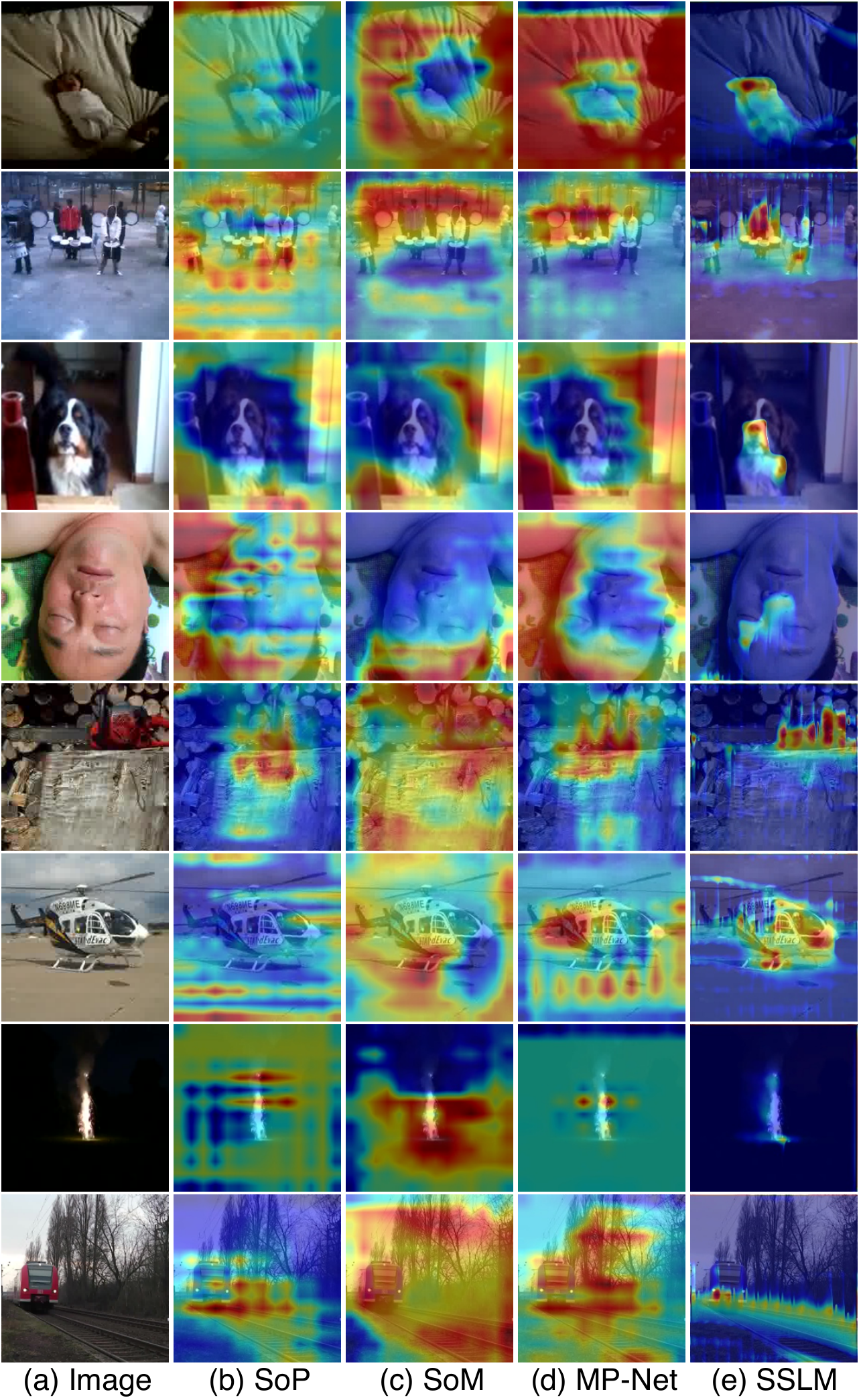}
    \caption{Visualizing sound source location of our proposed SSLM network in comparison with baseline methods SoP, SoM, and MP-Net on A-NATURAL dataset.}
    \label{fig:loc_anatural_vis}
\end{figure}

We visualize more examples of localized sounding sources by our proposed Sound Source Location Masking (SSLM) network in comparison with baseline methods of SoP~\cite{zhao2018sound}, SoM~\cite{zhao2019sound}, and MP-Net~\cite{xu2019recursive} on MUSIC, A-MUSIC and A-NATURAL datasets in Fig.~\ref{fig:loc_music_vis}, Fig.~\ref{fig:loc_amusic_vis}, and Fig.~\ref{fig:loc_anatural_vis} respectively.

\subsection{Datasets}

We evaluate the proposed approaches using Multimodal Sources of Instrument Combinations (MUSIC)~\cite{zhao2018sound} dataset, and two sub-sets of AudioSet~\cite{gemmeke2017audio}: A-MUSIC and A-NATURAL.

\subsubsection{MUSIC}
The MUSIC dataset is relatively small high quality dataset of musical instruments. It contains 714 untrimmed YouTube videos which span 11 instrumental categories, namely \textit{accordion}, \textit{acoustic guitar}, \textit{cello}, \textit{clarinet}, \textit{erhu}, \textit{flute}, \textit{saxophone}, \textit{trumpet}, \textit{tuba}, \textit{violin}, and \textit{xylophone}. For all the reported experiments, we randomly split the dataset into 400 training videos, 100 validation videos, and 130 test videos.

\subsubsection{A-MUSIC and A-NATURAL}
A-MUSIC dataset is a trimmed musical instrument dataset from AudioSet. It has around 25k videos spanning 10 instrumental categories: \textit{accordion}, \textit{bagpipe}, \textit{cello}, \textit{flute}, \textit{piano}, \textit{pizzicato}, \textit{saxophone}, \textit{trumpet}, \textit{ukulele}, and \textit{zither}. A-NATURAL dataset is a trimmed natural sound dataset from AudioSet. It contains around 10k videos which cover 10 categories of natural sounds, namely \textit{baby crying}, \textit{chainsaw}, \textit{dog}, \textit{drum}, \textit{firework}, \textit{helicopter}, \textit{printer}, \textit{rail}, \textit{snoring}, and \textit{water}. We split both the A-MUSIC and A-NATURAL dataset samples to 80\%, 10\%, and 10\% as train, validation and test set.

\subsection{Implementation Details}

\begin{figure}[!tbp]
    \centering
    \includegraphics[width=\textwidth,keepaspectratio]{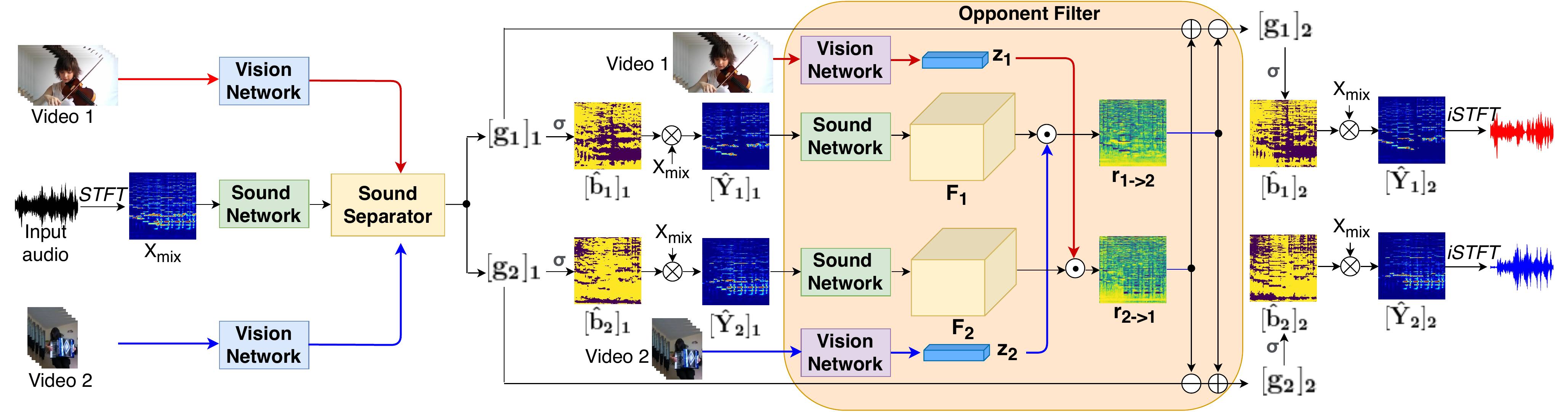}
    \caption{The overall architecture of the COF model in the case of two stages on two sound sources. In the first stage, visual representations (vision network) and sound features (sound network) are passed to the sound separator that produces the spectrum masks $g$ for each source. $g$ are then further provided as inputs for the upcoming stage. Stage two refines the separation result using visual representation $z_{2}$ to identify components $r_{1->2}$ from the source 1 that should belong to the source 2. The spectrum masks $g$ are updated accordingly by subtracting from $[g_{1}]_{1}$ and adding to $[g_{2}]_{1}$. Similar operation is done for the source 2. Finally, we obtain the separated audios by applying inverse STFT to the output component spectrograms $[\hat{Y}_{1}]_{2}$ and $[\hat{Y}_{2}]_{2}$, which are converted from $[g_{1}]_{2}$ and $[g_{2}]_{2}$ (Eq. (2) in main paper). Note that the vision networks at different stages can change accordingly to the vision network options discussed in Sec. 3.2 of main paper.}
    \label{fig:arch}
\end{figure}

\subsubsection{Overall Architecture}
We illustrate the overall architecture of the COF model in the case of two stages on two sound sources in Fig.~\ref{fig:arch}. The vision networks of the COF model at different stages change accordingly to the vision network options discussed in Sec. 3.2 of main paper. 

\subsubsection{Vision Network}
We extract video frames at 8fps and adopt frame augmentation by random scaling, random horizontal flipping, and random cropping ($224\times224$) during training for all datasets. We apply a dilated 2D ResNet18~\cite{he2016deep} with \textit{dilation}=2 to obtain representations of C2D-RGB and C2D-DYN. For a single input RGB image or dynamic image of size $\textit{3}\times\textit{16H}\times\textit{16W}$, we truncate the ResNet18 after \textit{stride}=16 and achieve the visual feature of size $\textit{K}\times\textit{H}\times\textit{W}$ by performing a 3 $\times$ 3 convolution with output channels of \textit{K}=16 on the top. The C3D models utilize 3D version of ResNet18 on $T$=48 frames. With the \textit{stride}=16 on spatial dimension and \textit{stride=8} on the temporal dimension, we yield the C3D-RGB and C3D-FLO representations of size $\textit{T}^{'}\times\textit{K}\times\textit{H}\times\textit{W}$, where $\textit{T}^{'}$=6 and $\textit{H}=\textit{W}=14$. 

\subsubsection{Mutual Attention Module}
The Mutual Attention (MA) module is proposed to fuse the appearance and motion information. In the MA module, we obtain the spatial attention map by projecting the appearance features from C2D-RGB to a single-channel feature map with a 1 $\times$ 1 convolution and a sigmoid operation. The MA module enhances the sound source relevant motions by multiplying the C3D features with the spatial attention map. The appearance-weighted features are added back to the original C3D features in order to keep C3D features as the principle cue in case the C2D-RGB fails to localize the sound source. We obtain C3D feature attention by adding a sigmoid function on top of the final enhanced C3D features. The multiplication between the C3D feature attention and the time-inflated appearance features are added back to the C2D-RGB appearance features. Within this process, for the predicted regions of interest from C2D-RGB, the appearance that has no motions will be eliminated. Finally, we receive the mutual attentive features of dimension ${\textit{T}}^{'}\times\textit{K}\times\textit{H}\times\textit{W}$ from the two-stream structures.

\subsubsection{Sound Network}
We adopt the U-Net~\cite{ronneberger2015u} with 7 layers of 2D CNN and output channels of \textit{K=16} as the architecture of Sound Network. The input audio signals are represented as spectrograms, which are obtained from the audio stream using Short-time Fourier transform (STFT). To obtain the final separated audio signals, the inverse STFT is applied to the component spectrograms.

\subsubsection{Sound Separator}

\begin{figure}[!tbp]
    \centering
    \includegraphics[width=0.9\textwidth,keepaspectratio]{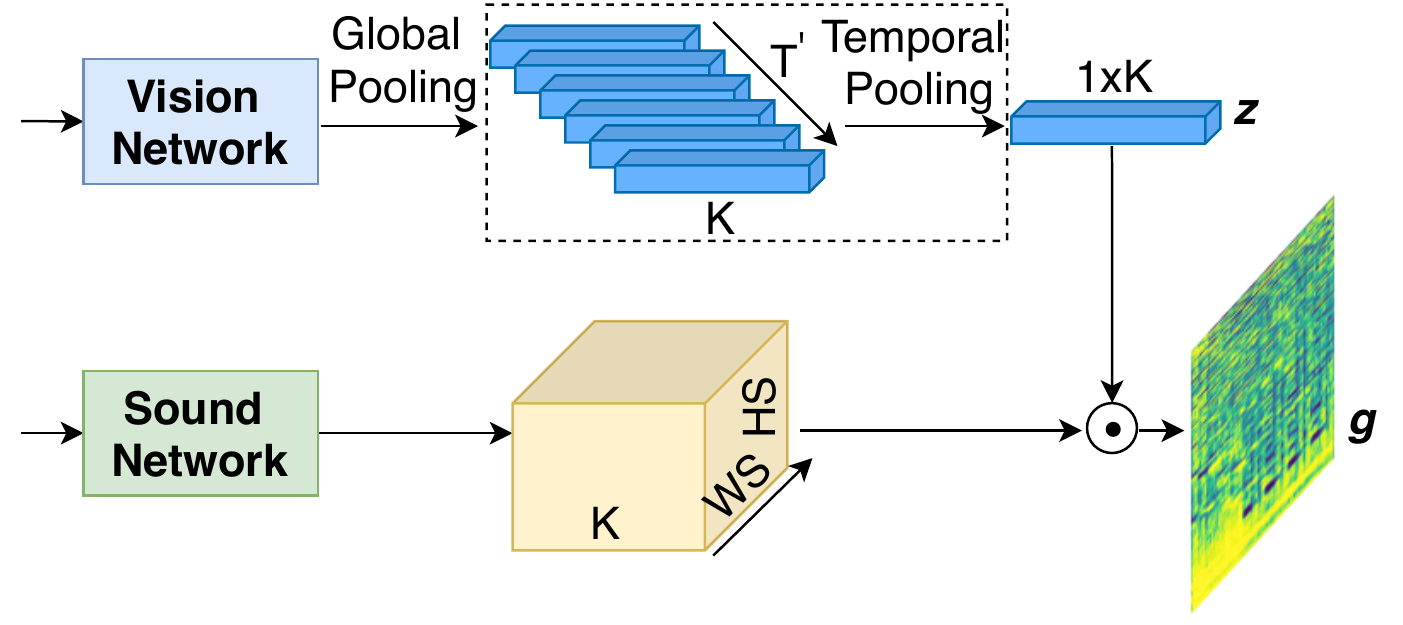}
    \caption{The architecture of sound separator. Sound separator combines visual representations $\mathbf{z}$ with sound network feature maps using a linear combination to predict the spectrum mask $g$. $g$ is then provided for the upcoming stage as input.}
    \label{fig:SS}
\end{figure}

We depict the architecture of the sound separator (Eq. (1) in main paper) in Fig.~\ref{fig:SS}. The sound separator combines the visual representations $\mathbf{z}$ with the sound network output using a linear combination to produce the spectrum mask $g$. Spectrum masks $g$ for all the sources are then provided for the upcoming stage as inputs.

\paragraph{\bf Optimization}

Our implementation is built on Pytorch. The network is trained with a batch size of 10 for 4,000 iterations. We use stochastic gradient descent (SGD) with momentum 0.9 and weight decay 1e-4 to train our Cascaded Opponent Filter (COF) network and Adam optimizer to train the Sound Source Location Masking (SSLM) network. The vision networks of COF and the SSLM, pre-trained on ImageNet~\cite{deng2009imagenet}, use a learning rate of 1e-4, while the rest of modules which are trained from scratch use a learning rate of 1e-3. We decrease the learning rate from its initial value by a factor of 10 every 1,600 iterations.

\end{document}